\documentclass{article} 
\usepackage{nips15submit_e,times}
\usepackage{hyperref}
\usepackage{url}

\usepackage{amsmath}
\usepackage{amsfonts}
\usepackage{amssymb}
\usepackage{mathtools}
\usepackage{graphicx}
\usepackage{caption}
\usepackage{subcaption}
\usepackage{verbatim}
\usepackage{algorithm}
\usepackage{algorithmic}

\title{Stochastic Collapsed Variational Inference for Sequential Data}

\author{
Pengyu Wang$^1$\ \ \ \ \  Phil Blunsom$^{1,2}$\\
$^1$Department of Computer Science, University of Oxford\\
$^2$Google DeepMind \\
\texttt{\{pengyu.wang,\,phil.blunsom\}@cs.ox.ac.uk}
}

\nipsfinalcopy 

\begin{document}

\maketitle

\begin{abstract}
Stochastic variational inference for collapsed models has recently been successfully applied to large scale topic modelling. In this paper, we propose a stochastic collapsed variational inference algorithm in the sequential data setting. Our algorithm is applicable to both finite hidden Markov models and hierarchical Dirichlet process hidden Markov models, and to any datasets generated by emission distributions in the exponential family. Our experiment results on two discrete datasets show that our inference is both more efficient and more accurate than its uncollapsed version, stochastic variational inference.
\end{abstract}

\section{Background}

A hidden Markov model (HMM) \cite{rabiner90hmm} consists of a hidden state sequence $\textbf{z} = \{z_t\}_{t=0}^T$ and a corresponding observation sequence $\textbf{x} = \{x_t\}_{t=1}^T$. Let there be $K$ hidden states. For convenience, we let the start state be $0$ and set $z_0 = 0$. Let $\theta$ be the transition matrix where $\theta_{k,k'}=p(z_t=k'|z_{t-1}=k)$, and $\theta_0$ be the initial state distribution where $\theta_{0,k'}=p(z_1=k')$. A hierarchical Dirichlet process HMM (HDP-HMM) \cite{beal02ihmm,teh06hdp} allows to use an unbounded number of hidden states by constructing an infinite mean vector $\pi$ from a stick breaking process and drawing transition vectors $\theta_k$ from the shared $\pi$. We have $\text{for } k=1,2,...$ and $\text{for } k'=1,2,...$,
\begin{align}
& \pi_{k'} = \tilde{\pi}_{k'}\prod_{l=1}^{k'-1}(1-\tilde{\pi}_l) & \tilde{\pi}_{k'} \sim \text{Beta}(1,\gamma) & &   \theta_{k} \sim \text{DP}(\alpha, \pi). & 
\end{align}

A hidden sequence is generated by a first order Markov process, and each observation is generated conditioned on its hidden state. We have for $t=1,...,T$,
\begin{align}
& z_t|z_{t-1}=k \sim \text{Mult}(\theta_{k}) &  x_t|z_t=k'  \sim p(\cdot|\phi_{k'}),
\end{align}
where $\phi_{k'}$ parametrizes the observation likelihoods for $k'=1,2,...$, with $\phi_{k',w}=p(x_t=w|z_t=k')$. We assume that the observation likelihoods and their conjugate prior take exponential forms:
\begin{align}
 p(w|\phi_{k'}) &= h_l(w) \exp \{ \phi_{k'}^T t(w) - a_l(\phi_{k'}) \}  \\
 p(\phi_{k'}|\lambda^\circ) &= h_g(\phi_{k'}) \exp \{ (\lambda_1^\circ)^T \phi_{k'} + (\lambda_2^\circ)^T (-a_l(\phi_{k'})) - a_g(\lambda^\circ) \} .
\end{align}

\begin{figure}
\centering
\includegraphics[scale=0.35]{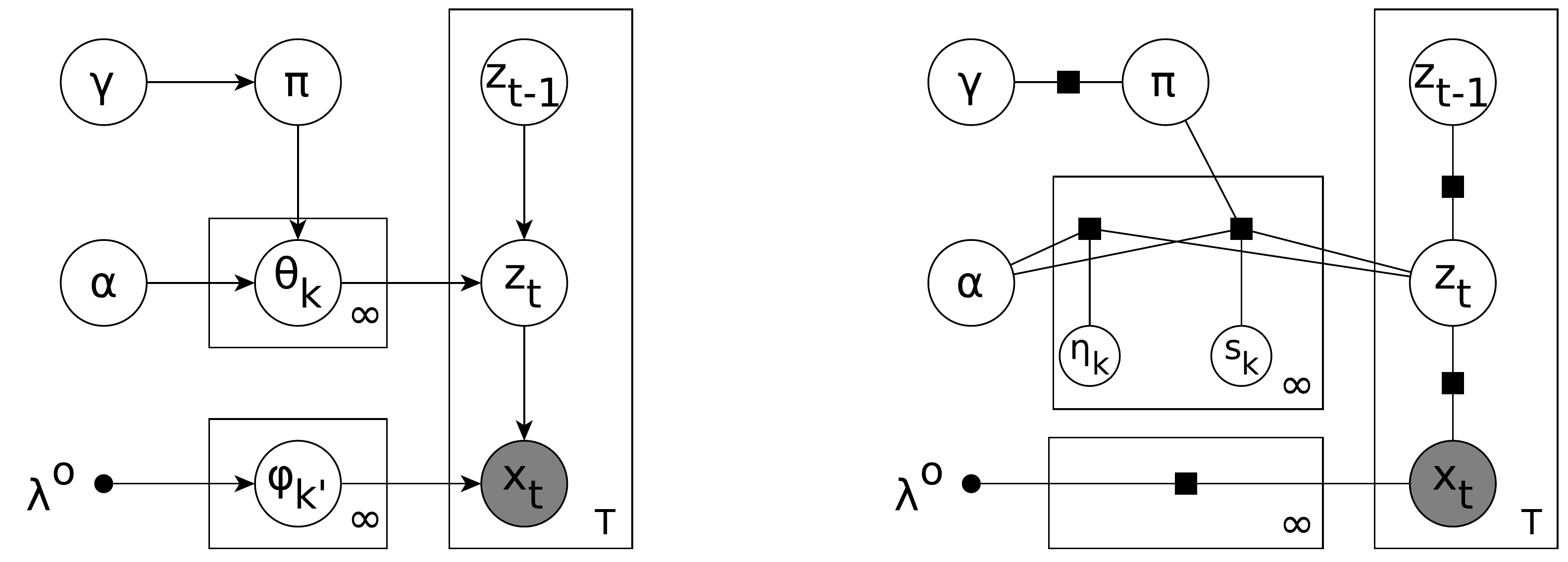}
\caption{Left: an HDP-HMM. Right: a collapsed HDP-HMM with auxiliary variables. Here we suppress the first order Markov dependencies into one plate repeated $T$ times to fit the page.}
\label{hdphmm}
\end{figure}

The base measure $h$ and log normalizer $a$ are scalar functions; and the parameter $\phi_{k'}$ and sufficient statistics $t$ are vector functions. The subscripts $l$ and $g$ represent the local hidden variables and global model parameters, respectively. The dimensionality of the prior hyperparameter $\lambda^\circ = (\lambda_1^\circ, \lambda_2^\circ)$ is equal to $\text{dim}(\phi_{k'})+1$. For a complete Bayesian treatment, we place vague Gamma priors on $\alpha$ and $\gamma$, $\alpha \sim \text{Gamma}(a^\circ_\alpha,b^\circ_\alpha)$ and $\gamma \sim \text{Gamma}(a^\circ_\gamma, b^\circ_\gamma)$. 
The graphical model is shown in figure \ref{hdphmm} (left).

\section{Stochastic Collapsed Variational Inference}

HMMs and HDP-HMMs are popular probabilistic models for modelling sequential data. However, their traditional inference methods such as variational inference (VI) \cite{beal03} and Markov chain Monte Carlo (MCMC) \cite{teh06hdp,vangael08beam} are not readily scalable to large datasets (e.g.,\ one dataset in our experiment contains $5$ million sequences with combined length over $100$ million). In this paper, we follow the success of stochastic collapsed variational inference (SCVI) for latent Dirichlet allocation (LDA) \cite{foulds13}, and we propose a scalable SCVI algorithm for HMMs and HDP-HMMs. Our algorithm achieved better predictive performances than the stochastic variational inference (SVI) \cite{hoffman13} applied to HMMs \cite{foti14} and to HDP-HMMs \cite{johnson14}.

We present our derivation in the following three steps: 1.\ we marginalize out the model parameters $(\theta,\phi)$; 2.\ we derive stochastic updates for each sequence; 3.\ we derive stochastic updates for the posteriors of the HDP parameters $(\alpha,\tilde{\pi},\gamma)$ (for HDP-HMMs only). For notational simplicity, we consider a dataset of $N$ sequences each of length $T$. That is $\textbf{x} = \{\textbf{x}^n\}_{n=1}^N$ and $\textbf{x}^n=\{x^n_t\}_{t=1}^T$. Similarly, we write for hidden sequences $\textbf{z} = \{\textbf{z}^n\}_{n=1}^N$ and $\textbf{z}^n=\{z^n_t\}_{t=1}^T$.

\subsection{Collapsed HDP-HMMs}

There is substantial empirical evidence \cite{asuncion09,wangandblunsom13,foulds13} that marginalizing the model parameters is helpful for both accurate and efficient inference. The marginal data likelihood of an HDP-HMM is:
\begin{align}
 p(\textbf{x},\textbf{z}) = \prod_{k=0}^K \textstyle \frac{\Gamma(\alpha)}{\Gamma(\alpha+C_{k\cdot})} \prod_{k'=1}^K \frac{\Gamma(\alpha \pi_{k'} + C_{kk'})}{\Gamma(\alpha \pi_{k'})}
 \displaystyle \prod_{n=1}^N \prod_{t=1}^T \textstyle h_l(x^n_t)  \displaystyle \prod_{k'=1}^K \textstyle \exp \{ a_g(\lambda^{k'}) -  \{ a_g(\lambda^\circ)\}.
 \label{collapsedhdphmm}
\end{align}
The gamma functions and log normalizers come from the marginalization. $C_{kk'}$ denotes the transition count from the hidden state $k$ to $k'$, $C_{kk'} = \# \{n,t: z^n_{t-1}=k,z^n_t={k'}\}$. dot denotes the summed out column, e.g.,\ $C_{\cdot k'}= \sum_{k} C_{kk'}$. $\lambda^{k'}$ denotes the posterior hyperparameter for the hidden state $k'$, $\lambda^{k'}_1 = \lambda^\circ_1 + \sum_{n=1}^N \sum_{t=1}^T t(x^n_t)\delta(z^n_t=k')$ and $\lambda^{k'}_2 =\lambda^\circ_2 + C_{\cdot k'}$, where $\delta$ is the standard delta function.

The gamma functions are a nuisance to take derivatives of (\ref{collapsedhdphmm}). Following \cite{teh08}, we replace them by integrals of some auxiliary variables $\eta$ and $\textbf{s}$ and the joint likelihood becomes:
\begin{align}
&p(\textbf{x},\textbf{z},\eta,\textbf{s}) \notag \\
= &\prod_{k=0}^K \textstyle \frac{\eta_k^{\alpha-1} (1-\eta_k)^{C_{k\cdot}-1} } {\Gamma(C_{k\cdot})} \prod_{k'=1}^K {C_{kk'} \brack s_{kk'}} (\alpha\pi_{k'})^{s_{kk'}} 
\displaystyle \prod_{n=1}^N \prod_{t=1}^T  \textstyle h_l(x^n_t)  \displaystyle \prod_{k'=1}^K \textstyle \exp \{ a_g(\lambda^{k'})-\{ a_g(\lambda^\circ)\}
\label{expandedhdphmm}
\end{align}
where $\eta_k \in [0,1]$ is Beta distributed, $s_{kk'} \in \{0,1,...,C_{kk'}\}$ is the number of tables labelled with $k'$ in the $k^{\text{th}}$ Chinese restaurant in a Chinese restaurant franchise, and ${C_{kk'} \brack s_{kk'}}$ is unsigned Stirling number of the first kind. The factor graph with the auxiliary variables is given in figure \ref{hdphmm} (right).

We are interested in the posterior $p(\textbf{z},\eta,\textbf{s},\gamma,\alpha,\tilde{\pi}|\textbf{x})$. As the exact computation is intractable, we introduce a variational distribution in a tractable family,
\begin{align}
q(\textbf{z},\eta,\textbf{s},\gamma,\alpha,\tilde{\pi}) = q(\textbf{z}) q(\eta|\textbf{z})q(\textbf{s}|\textbf{z})q(\gamma)q(\alpha)q(\tilde{\pi}),
\end{align}
and we maximize the evidence lower bound (ELBO) denoted by $\mathcal{L}(q)$,
\begin{align}
\log p(\textbf{x}) \geq \mathbb{E}[\log p(\textbf{x},\textbf{z},\eta,\textbf{s},\gamma,\alpha,\tilde{\pi})] 
- \mathbb{E}[\log q(\textbf{z},\eta,\textbf{s},\gamma,\alpha,\tilde{\pi}) ] \triangleq \mathcal{L}(q).
\label{ELBO}
\end{align}

By the `direct assignment truncation' \cite{teh08,johnson14}, we set the truncation level to be $K$. That is $q(\textbf{z}=0)$ if for any $n$ and $t$ such that $z^n_t>K$.

\subsection{Inference for Sequences}

To infer $q(\textbf{z})$, we factorize it as a product of independent sequences, $q(\textbf{z}) = \prod_{n=1}^N q(\textbf{z}^n)$. Combining the work of SCVI for LDA \cite{foulds13} and CVI for HMM \cite{wangandblunsom13}, we randomly sample $\textbf{x}^n$ with $n\sim \mathcal{U}[1,N]$, and we derive the update for $q(\textbf{z}^n)$ with a zeroth order Taylor approximation \cite{teh07collapsed}:
\begin{align}
 q(\textbf{z}^n)  &\propto \prod_{t=1}^{T} \hat{\theta}_{z^n_{t-1},z^n_t} \prod_{t=1}^T \hat{\phi}_{z^n_t,x^n_t} \quad\quad\quad
 \hat{\theta}_{k,k'}  \propto \mathbb{G}[\alpha \pi_{k'}]+\mathbb{E}[C_{kk'}]  \label{scvihmm1-1} \\
  \hat{\phi}_{k',w} &\propto h(w) \exp \{ a_g( \lambda^\circ_1+t(w)+\mathbb{E}[t_{k'}(\textbf{x},\textbf{z})],  \lambda^\circ_2+1+\mathbb{E}[C_{\cdot k'}]) \},
\label{scvihmm1-2}
\end{align}
in which $\mathbb{G}$ denotes the geometric expectation, $\mathbb{E}[C_{kk'}]$ denotes the expected transition count from state $k$ to $k'$, and $\mathbb{E}[t_{k'}(\textbf{x},\textbf{z})] = \sum_{n=1}^N \sum_{t=1}^T q(z^n_t=k')t(x^n_t)$ denotes the emission statistics at the hidden state $k'$. The details on expectations that appear in the paper are in Appendix A.

As $q(\textbf{z}^n)$ is proportional to a HMM parametrized by the surrogate parameters $\hat{\theta}$ and $\hat{\phi}$, we can use the forward backward algorithm \cite{baum66}. After collecting the local transition counts $\mathbb{E}[C^n_{kk'}]$ and emission statistics $\mathbb{E}[t_{k'}(\textbf{x}^n,\textbf{z}^n)]$, we update the global statistics by taking a weighted average:
\begin{align}
\mathbb{E}[C_{kk'}] &= (1-\rho_n) \mathbb{E}[C_{kk'}] + \rho_n N \mathbb{E}[C^n_{kk'}] \label{scvihmm2-1}\\
\mathbb{E}[t_{k'}(\textbf{x},\textbf{z})] &= (1-\rho_n) \mathbb{E}[t_{k'}(\textbf{x},\textbf{z})] + \rho_n N \mathbb{E}[t_{k'}(\textbf{x}^n,\textbf{z}^n)],
\label{scvihmm2-2}
\end{align}
where $\rho_n$ is the step size satisfying $\sum_n \rho_n^2 \leq \infty$ and $\sum_n \rho_n = \infty$.

Unlike CVI for HMM \cite{wangandblunsom13}, our algorithm is memory efficient, since we update $q(\textbf{z}^n)$ without subtracting the local statistics, as such they do not need to be explicitly stored.

\subsection{Inference for HDP}

For notational clarity, we write the variational posteriors of the HDP parameters to be governed by their variational parameters. We have,
\begin{align}
& q(\tilde{\pi}_{k'}) = \text{Beta}(u_{k'},v_{k'}) & q(\alpha) = \text{Gamma}(a_\alpha,b_\alpha) && q(\gamma) = \text{Gamma}(a_\gamma,b_\gamma). &
\end{align}

We derive stochastic updates for the HDP posteriors. For a randomly selected sequence $\textbf{x}^n$, we form an artificial dataset $\{\textbf{x}^{n^{(N)}},\textbf{z}^{n^{(N)}}\}$ consisting $N$ replicates of the observed and hidden sequences $\{\textbf{x}^n,\textbf{z}^n\}$. Assuming we can compute $\mathbb{E}[s_{kk'}^{(N)}]$ and $\mathbb{E}[\log \eta_k^{(N)}]$ based on the artificial dataset, we derive the intermediate variational parameters and take a weighted average with their old estimates. Hence, we have the following updates ($\mathbb{E}[\log (1-\tilde{\pi}_{k'})]$ in (\ref{scvihdp6}) is also a function of $\mathbb{E}[s_{kk'}^{(N)}]$):
\begin{align}
\! \! u_{k'} &= (1-\rho_n) u_{k'} +  \rho_n (1 + \mathbb{E}[s^{(N)}_{\cdot k'}])
&\! v_{k'} &= (1-\rho_n) v_{k'} + \rho_n (\mathbb{E}[\gamma] + \mathbb{E}[s^{(N)}_{\cdot >k'}])  \label{scvihdp2} \\
\! \! a_\alpha &= (1-\rho_n)a_\alpha + \rho_n(a^\circ_\alpha + \mathbb{E}[s^{(N)}_{\cdot\cdot}])
&\! b_\alpha &= (1-\rho_n)b_\alpha + \rho_n (b^\circ_\alpha - \textstyle \sum_{k} \mathbb{E}[\log \eta^{(N)}_k])  \label{scvihdp4} \\
\! \! a_\gamma &= (1-\rho_n)a_\gamma + \rho_n (a^\circ_\gamma + K)
&\! b_\gamma &= (1-\rho_n)b_\gamma + \rho_n (b^\circ_\gamma -  \textstyle \sum_{k'} \mathbb{E}[\log (1-\tilde{\pi}_{k'})] ) \label{scvihdp6}
\end{align}
where dot denotes the summed out column, and $>k'$ denotes summing over $l$ for $l>k'$. The details on computing $\mathbb{E}[s_{kk'}^{(N)}]$ and $\mathbb{E}[\log \eta_k^{(N)}]$ are in Appendix B. Stochastic optimizations often benefit from the use of minibatches, to reduce the variance of noisy samples and the updating time of variational parameters. Thus we propose to update the global statistics after a minibatch is processed, and to update the HDP posteriors after a larger batch is processed. Altogether, our SCVI algorithm for HDP-HMMs is given in Appendix C, and it applies to HMMs by removing the outermost loop.

\begin{figure}[t]
\begin{center}
\includegraphics[scale=0.3795]{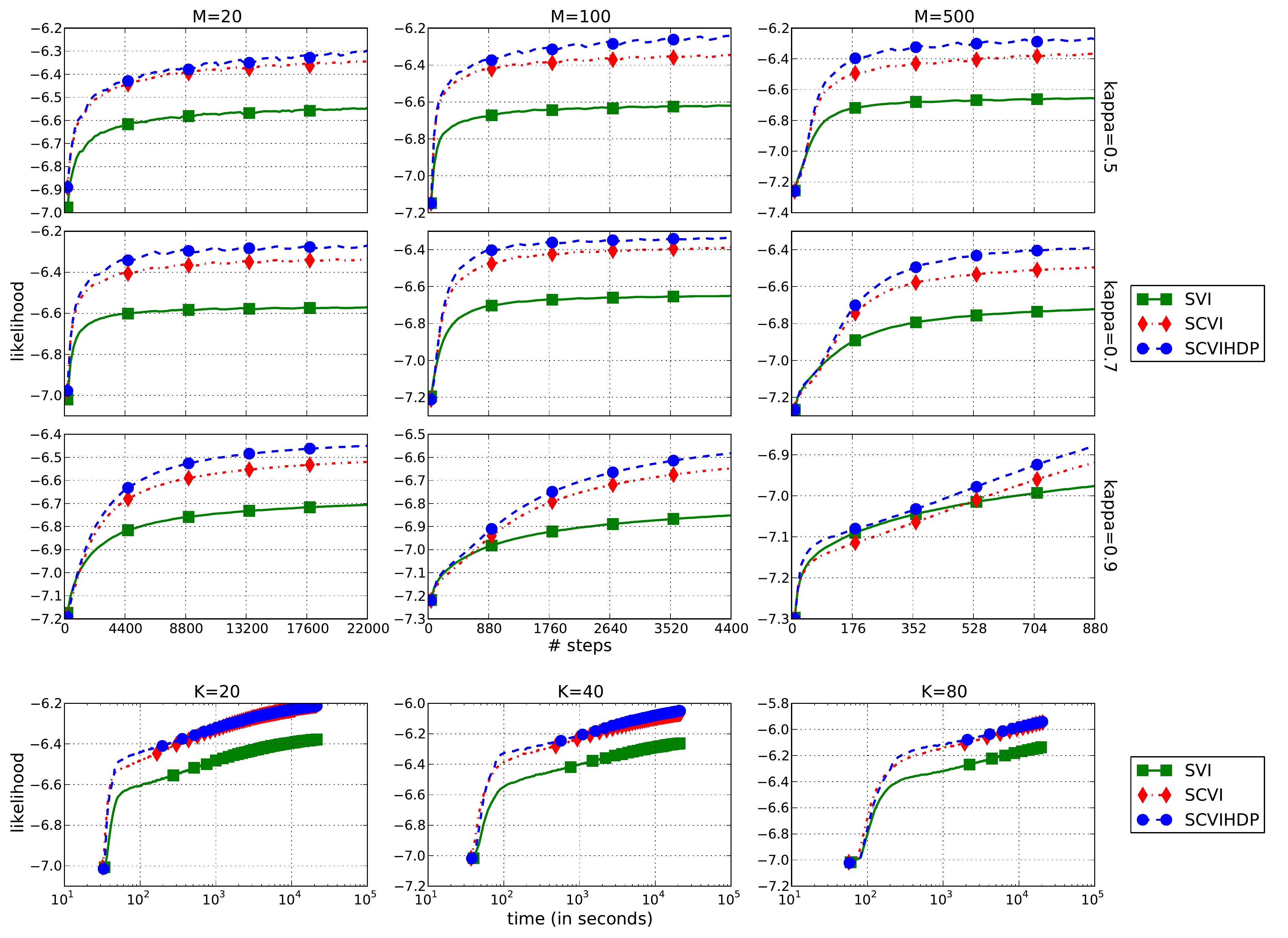}
\caption{Top three rows: comparison on WSJ under various minibatch sizes $M$ and forgetting rates $\kappa$. Bottom row: comparison on NYT under various (truncated) numbers of hidden states $K$.}
\label{combined_online}
\end{center}
\end{figure}

\section{Experiments}

We evaluated our SCVI algorithm applied to HMMs (denoted by SCVI) and applied to HDP-HMMs (denoted by SCVIHDP) compared to the SVI algorithm applied to HMMs \cite{foti14} (denoted by SVI). SVI applied to HDP-HMMs was omitted, since we were unable to make noticeable improvement over SVI using a point estimate of the top level stick $\pi$ \cite{johnson14}. We used two discrete datasets, the Wall Street Journal (WSJ) and New York Times (NYT). Both datasets are made of sentences. For each sentence, the underlying sequence can be understood as a Markov chain of hidden part-of-speech (PoS) tags \cite{jurafsky00} and words are drawn conditioned on PoS tags, making (HDP)-HMMs natural models. We used the predictive log likelihoods as our evaluation metrics.

For SVI and SCVI, we set the transition priors to $\text{Dir}(0.1)$, to encourage sparsity. For SCVIHDP, we set the HDP priors to be vague, $\text{Gamma}(1,0.1)$. We set $\mathbb{G}[\alpha \pi_{k'}]=0.1$ for the first iteration such that all the algorithms started with the same transition prior counts. Finally, all the emission priors were set to $\text{Dir}(0.1)$; all the global statistics $\mathbb{E}[C_{k,k'}]$ and $\mathbb{E}[t_{k'}(\textbf{x},\textbf{z})]$ were initialized using exponential distributions, as suggested by \cite{hoffman13}.

The first three rows in figure \ref{combined_online} presents the predictive log likelihood results of three inferences on WSJ ($49,000$ sentences, $90\%$ for training and $10\%$ for testing). We fixed the number of hidden states (or truncation level) $K=45$\footnote{One goal of our experiments is to show the improvement by sharing statistics using our HDP inference. Using a larger truncation level than the number of hidden states would put our SCVIHDP in an advantageous position and we would not be able to identify the source of improvement.} and varied the minibatch sizes $M$ and forgetting rates $\kappa$, which parametrize the step sizes $\rho_n = (1+n)^{-\kappa}$. The large batch size was set to be $10,000$ for SCVIHDP. We let each inference run through the dataset $10$ times and reported the per time step likelihoods. In all the settings, our SCVI outperformed SVI by large margins, extending the success of SCVI for LDA \cite{foulds13} to time series data. Further, our collapsed HDP inference helped SCVIHDP to surpass our SCVI by noticeable margins.

The forth row in figure \ref{combined_online} presents the predictive log likelihood results of three inferences on NYT ($5$ million sentences, $99\%$ for training and $1\%$ for testing) using the complementary settings to WSJ. We fixed $\kappa=0.5$ and $M=1000$ and varied $K$. We ran all the algorithms (implemented in Cython) for $6$ hours and reported the likelihood results versus wall-clock time. We see that given the same time, our SCVI converged much better than SVI. Our SCVIHDP overlapped with our SCVI towards the end, but it was always better prior to that, making better use of its time.

\section{Conclusion}

In this paper, we have presented a general stochastic collapsed variational inference algorithm that is scalable to very large time series datasets, memory efficient and significantly more accurate than the existing SVI algorithm. Our algorithm is also the first truly variational algorithm for HDP-HMMs, avoiding point estimates, and it comes with performance gains. For future work, we aim to derive the true nature gradients of the ELBO to prove and further speed up the convergence of our algorithm \cite{ruiz2014}, although we never saw a nonconverging case in our experiments.

\small{
\bibliographystyle{unsrt}
\bibliography{standard}
}

\clearpage

\normalsize

\section*{Appendix A}

In this section, we present the standard (geometric) expectations that appeared in the main paper.

If $x$ is Beta distributed, $p(x|u,v) \propto x^{u-1} (1-x)^{v-1}$, (e.g.,\ $\tilde{\pi}_{k'}$ in the main paper), we have,
\begin{align}
& \mathbb{E}[\log x] = \psi(u)-\psi(u+v) & \mathbb{E}[\log (1-x)] = \psi(v)-\psi(u+v) &
\end{align}
If $x$ is Gamma distributed, $p(x|a,b) \propto x^{a-1} e^{-bx}$ (e.g.,\ $\alpha$ in the main paper), we have,
\begin{align}
& \mathbb{E}[x] = a/b & \mathbb{G}[x] = e^{\psi(a)}/b &
\end{align}
If $x$ and $y$ are independent, (e.g.,\ $\alpha$ and $\pi_{k'}$ in the main paper), we have $\mathbb{G}[xy] = \mathbb{G}[x]\mathbb{G}[y]$.

\section*{Appendix B}

In this section, we present the details on computing $\mathbb{E}[s_{kk'}^{(N)}]$ and $\mathbb{E}[\log \eta_k^{(N)}]$. For $\mathbb{E}[s_{kk'}^{(N)}]$, we notice the inequality\footnote{ $\mathbb{E}[s^{n}_{k k'}]$ is the expected number of tables $k'$ in the $k^{th}$ restaurant. The inequality holds by the property of CRP: the expected number of tables grows not linearly but logarithmically with the number of customers.}: $\mathbb{E}[s_{kk'}^{(N)}] \neq N\mathbb{E}[s_{kk'}^{n}]$. Thus we compute it as follows:
\begin{align}
\mathbb{E}[s^{(N)}_{kk'}] &\approx \mathbb{G}[\alpha\pi_{k'}]q(C^{(N)}_{kk'}>0) (\psi(\mathbb{G}[\alpha\pi_{k'}] + \mathbb{E}_+[C^{(N)}_{kk'}]) - \psi(\mathbb{G}[\alpha\pi_{k'}])) \label{s_1} \\
q(C^{(N)}_{kk'}>0) &= 1-q(C^{(N)}_{kk'}=0) = 1- \exp \{ N \log q(C^n_{kk'}=0) \} \label{s_2} \\
q(C^n_{kk'}=0) &\approx \exp \{\textstyle \sum_t \log (1 - q((z^n_{t-1},z^n_t) = (k,k')) ) \} \label{s_3} \\
\mathbb{E}_+[C^{(N)}_{kk'}] &\approx N\mathbb{E}[C^{n}_{kk'}]/q(C^{(N)}_{kk'}>0) \label{s_4}
\end{align}
The approximation in (\ref{s_1}) comes from the technique proposed by Teh et al.\ detailed in \cite{teh08}. In (\ref{s_2}), $q(C^{(N)}_{kk'}>0)$ denotes the probability of at least one transition from state $k$ to state $k'$; and the second equality comes from the fact that $\textbf{z}^n$ is repeated $N$ times under exactly the same distribution. In (\ref{s_3}) and (\ref{s_4}), we propose a fast approximate method. We partition a hidden sequence $\textbf{z}^n$ into a set of overlapping but independent clusters $\{(z^n_t,z^n_{t+1})\}_{t=1}^T$. Allowing to overlap is sufficient to preserve all the pairwise transition information, while making the independence assumption permits the above linear computations as in \cite{teh08}. The same strategy applies to computing $\mathbb{E}[\log \eta_k^{(N)}]$.
\begin{align}
 \mathbb{E}[\log \eta_k^{(N)}] &\approx q(C^{(N)}_{k\cdot}>0) (\psi(\mathbb{E}[\alpha]) - \psi(\mathbb{E}[\alpha] + \mathbb{E}_+[C^{(N)}_{k\cdot}])) \\
 q(C^{(N)}_{k\cdot}>0) &= 1-q(C^{(N)}_{k\cdot}=0) = 1- \exp \{ N \log q(C^n_{k\cdot}=0) \} \label{eta_2} \\
q(C^n_{k\cdot}=0) &\approx \exp \{\textstyle \sum_t \log (1 - q((z^n_{t-1}) = (k)) ) \} \label{eta_3} \\
\mathbb{E}_+[C^{(N)}_{k\cdot}] &\approx N\mathbb{E}[C^{n}_{k\cdot}]/q(C^{(N)}_{k\cdot}>0) \label{eta_4}
\end{align}

\section*{Appendix C}
In this section, we present our SCVI algorithm for HDP-HMMs.
\begin{algorithm}[h]                    	
\caption{SCVI for HDP-HMMs (and for HMMs by deleting the outermost loop)}          	
\label{algo}                          	
\begin{algorithmic}                    	
	\STATE Randomly initialize $\mathbb{E}[C_{k,k'}]$ and $\mathbb{E}[t_{k'}(\textbf{x},\textbf{z})]$
	\FOR{each large batch}
    	\FOR{each mini batch}
    		\FOR{each sequence $\textbf{x}^n$}
				\STATE update $q(\textbf{z}^n)$ by Eqs. (\ref{scvihmm1-1},\ref{scvihmm1-2})
			\ENDFOR
			\STATE update $\mathbb{E}[C_{k,k'}],\mathbb{E}[t_{k'}(\textbf{x},\textbf{z})]$  by Eqs. (\ref{scvihmm2-1},\ref{scvihmm2-2})
		\ENDFOR
        \STATE update $q(\pi),q(\alpha),q(\gamma)$ by Eqs. (\ref{scvihdp2},\ref{scvihdp4},\ref{scvihdp6})
    \ENDFOR
\end{algorithmic}
\end{algorithm}

\end{document}